\documentclass[sigconf]{acmart}

\AtBeginDocument{%
  \providecommand\BibTeX{{%
    \normalfont B\kern-0.5em{\scshape i\kern-0.25em b}\kern-0.8em\TeX}}}

\setcopyright{acmcopyright}
\copyrightyear{2018}
\acmYear{2018}
\acmDOI{10.1145/1122445.1122456}

\acmConference[Woodstock '18]{Woodstock '18: ACM Symposium on Neural
  Gaze Detection}{June 03--05, 2018}{Woodstock, NY}
\acmBooktitle{Woodstock '18: ACM Symposium on Neural Gaze Detection,
  June 03--05, 2018, Woodstock, NY}
\acmPrice{15.00}
\acmISBN{978-1-4503-XXXX-X/18/06}

\copyrightyear{2020} 
\acmYear{2020} 
\setcopyright{acmcopyright}\acmConference[ICAIF '20]{ACM International Conference on AI in Finance}{October 15--16, 2020}{New York, NY, USA}
\acmBooktitle{ACM International Conference on AI in Finance (ICAIF '20), October 15--16, 2020, New York, NY, USA}
\acmPrice{15.00}
\acmDOI{10.1145/3383455.3422544}
\acmISBN{978-1-4503-7584-9/20/10}

\begin{document}

\title{Trading via Image Classification}

\author{Naftali Cohen}
\authornotemark[1]
\affiliation{%
  \institution{J.~P.~Morgan AI Research}
  \city{New York}
  \state{NY}}
\email{naftali.cohen@jpmchase.com}

\author{Tucker Balch}
\affiliation{%
  \institution{J.~P.~Morgan AI Research}
  \city{New York}
  \state{NY}}
\email{tucker.balch@jpmchase.com}

\author{Manuela Veloso}
\affiliation{%
  \institution{J.~P.~Morgan AI Research}
  \city{New York}
  \state{NY}}
\email{manuela.veloso@jpmchase.com}

\renewcommand{\shortauthors}{Cohen N., et al.}

\begin{abstract}
The art of systematic financial trading evolved with an array of approaches, ranging from simple strategies to complex algorithms, all relying primarily on aspects of time-series analysis (e.g., \citeauthor{murphy1999technical}, \citeyear{murphy1999technical}; \citeauthor{de2018advances}, \citeyear{de2018advances}; \citeauthor{tsay2005analysis}, \citeyear{tsay2005analysis}).
After visiting the trading floor of a leading financial institution, we noticed that traders always execute their trade orders while \emph{observing} images of financial time-series on their screens. In this work, we build upon image recognition's success (e.g., \citeauthor{krizhevsky2012imagenet}, \citeyear{krizhevsky2012imagenet}; \citeauthor{szegedy2015going}, \citeyear{szegedy2015going};
\citeauthor{zeiler2014visualizing}, \citeyear{zeiler2014visualizing}; \citeauthor{wang2017residual}, \citeyear{wang2017residual}; \citeauthor{koch2015siamese}, \citeyear{koch2015siamese}; \citeauthor{lecun2015deep}, \citeyear{lecun2015deep}) and examine the value of transforming the traditional time-series analysis to that of image classification. 
We create a large sample of financial time-series images encoded as candlestick (Box and Whisker) charts and label the samples following three algebraically-defined binary trade strategies (\citeauthor{murphy1999technical}, \citeyear{murphy1999technical}). Using the images, we train over a dozen machine-learning classification models and find that the algorithms efficiently recover the complicated, multiscale label-generating rules when the data is visually represented.   
We suggest that the transformation of continuous numeric time-series classification problem to a vision problem is useful for recovering signals typical of technical analysis.
\end{abstract}



\maketitle

\section{Introduction}\label{sec:introduction}
Traders in the financial markets execute buy and sell orders of financial instruments as stocks, mutual funds, bonds, and options daily. They execute orders while reading news reports and earning calls. Concurrently, they observe charts of time-series data that indicates the historical value of securities and leading financial indices (see Fig.~1 for a typical workstation of a professional trader\footnote{The photo was taken in a trading room at Rouen, Normandie, France, September 2015.}). Many algorithms have been developed to analyze continuous financial time-series data to improve a trader's decision-making ability to buy or sell a particular security (\citeauthor{murphy1999technical}, \citeyear{murphy1999technical}).  Conventional algorithms process time-series data as a list of numerical data, aiming at detecting patterns as trends, cycles, correlations, etc. (e.g., \citeauthor{de2018advances}, \citeyear{de2018advances}; \citeauthor{tsay2005analysis}, \citeyear{tsay2005analysis}). If a pattern is identified, the analyst can then construct an algorithm that will use the detected pattern (e.g., \citeauthor{wilks2011statistical}, \citeyear{wilks2011statistical}) to predict the expected future values of the sequence at hand (i.e., forecasting using exponential smoothing models, etc.).

Experienced traders, who observe financial time-series charts and execute buy and sell orders, start developing an intuition for market opportunities. The intuition they develop based on their chart observations nearly reflects the recommendations that their state-of-the-art model provides (personal communication with J.P.~Morgan's financial experts Jason Hunter, Joshua Younger, Alix Floman, and Veronica Bustamante).
In this perspective, financial time-series analysis can be thought of as a visual process. That is, when experienced traders look at a time-series data, they process and act upon the image instead of mentally exercising algebraic operations on the sequence of numbers. 

\begin{figure}[htb]
\centering
\noindent\includegraphics[width=1\columnwidth]{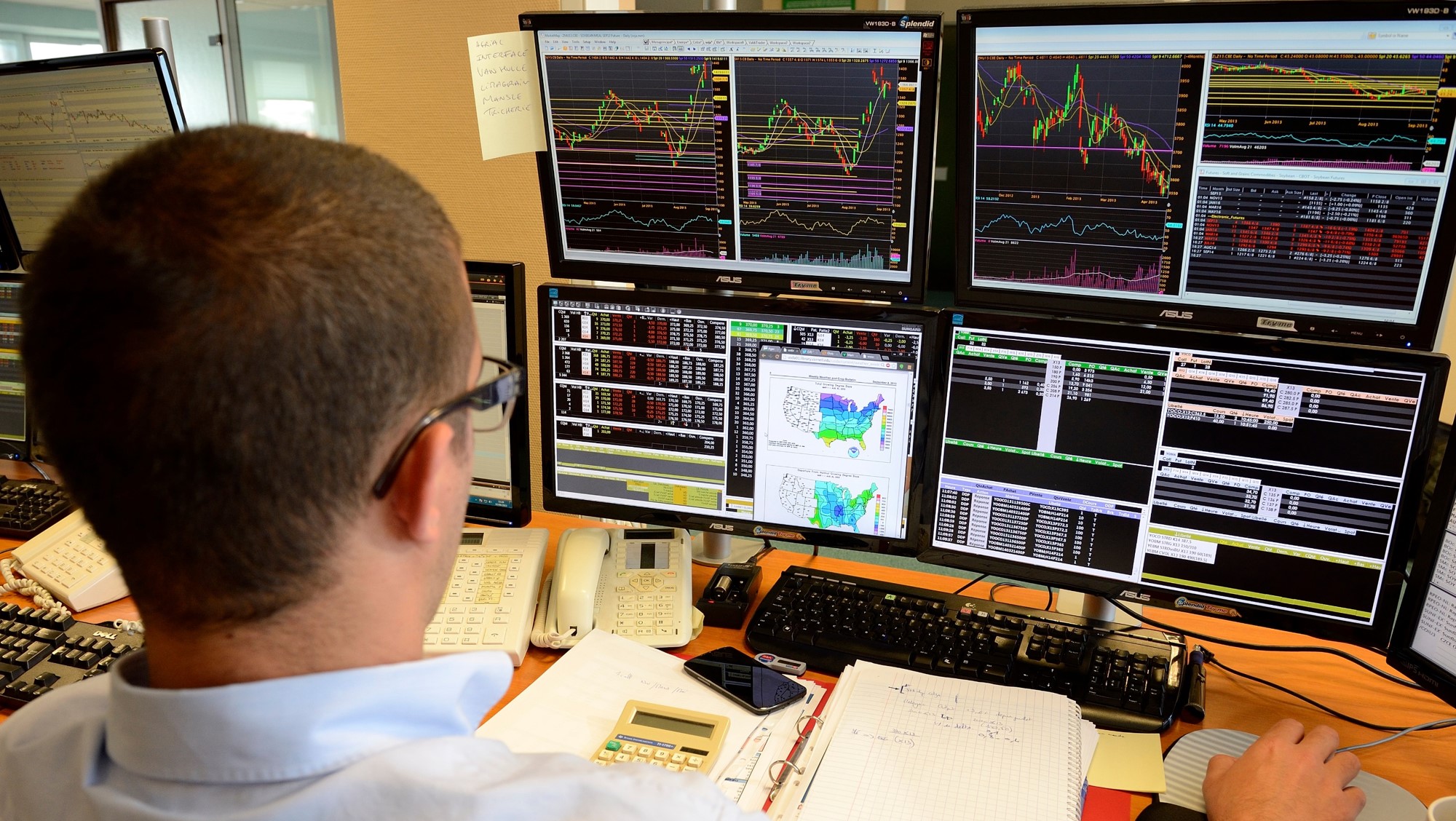}
\caption{Typical workstation of a professional trader. Credit: Photoagriculture / Shutterstock.com.}
\end{figure}

This paper is written under the assumption that, given public knowledge, markets are efficient (e.g., \citeauthor{pedersen2019efficiently}, \citeyear{pedersen2019efficiently}). That is, future market movements have almost no predictability. However, the way professionals trade is systematic (e.g., consistent, back-tested, and potentially profitable for a short duration) and can be characterized using a set of rules. 
We ask, can we build a system that identifies and replicates the way humans trade? For this, we create extensive financial time-series image data.
We make use of three known label-generating rules following algebraically-defined binary trade strategies (\citeauthor{murphy1999technical}, \citeyear{murphy1999technical}) to replicate the way people trade.
Using a supervised classification approach (e.g., \citeauthor{bishop2006pattern}, \citeyear{bishop2006pattern}; \citeauthor{goodfellow2016deep}, \citeyear{goodfellow2016deep}; \citeauthor{aggarwal2015data}, \citeyear{aggarwal2015data}), we evaluate predictions using over 15 different classifiers and show that the models are very efficient in identifying the complicated, sometimes multiscale, labels.

\section{Related Work and Main Contributions}
The focus of this work is on the representation of financial time-series data as images.
Previous work on time-series classification suggests first transforming the data either locally using wavelets or globally using Fourier transforms and then compare the modes of variability in the transformed spaces (e.g., \citeauthor{wilks2011statistical}, \citeyear{wilks2011statistical}). Other methods apply similarity metrics such as Euclidean distance, k-nearest neighbors, dynamic time warping, or even Pearson correlations to separate the classes (e.g., \citeauthor{aggarwal2015data}, \citeyear{aggarwal2015data}). 
In addition to the above, other techniques focus on manual feature engineering to detect a frequently occurring pattern or shape in the time series (e.g., \citeauthor{bagnall2017great}, \citeyear{bagnall2017great}). 

More recently, it was suggested to approach time-series classification by first encoding the data as images and then utilize the power of computer vision algorithms for classification (\citeauthor{park2019specaugment}, \citeyear{park2019specaugment}). In an example, it was suggested to encode the time dependency, implicitly, as Gramian-Angular fields, Markov-Transition fields (\citeauthor{wang2015encoding}, \citeyear{wang2015encoding}; \citeauthor{wang2015imaging}, \citeyear{wang2015imaging}), or make use of recurrence plots (\citeauthor{souza2014extracting}, \citeyear{souza2014extracting}; \citeauthor{silva2013time}, \citeyear{silva2013time}; \citeauthor{hatami2018classification}, \citeyear{hatami2018classification}) as a graphical representation.
Another work focused on transforming financial data into images to classify candlesticks patterns (\citeauthor{tsai2019encoding}, \citeyear{tsai2019encoding}).

In this paper, we examine the value of images alone for identifying trade opportunities typical for technical analysis. To the best of our knowledge, our work is the first that built upon the great success in image recognition and tries to systematically apply it to numeric time-series classification by taking a \emph{direct} graphical approach and recency-biased label-generating rules. 
The contributions of this paper are as follows:
\begin{enumerate}
\item The first contribution is bridging the unrelated areas of quantitative finance and computer vision. The former involves a mixture of technical, quantitative analysis, and financial knowledge, while the second involves advanced algorithm design and computer science techniques. In this paper, we show how the two distinct areas can leverage expertise and methods from each other.

\item The second contribution is our understanding that, in practice, there are financial domains in which investment decisions are made using visual representations alone (e.g., swap trade) -- relying, in fact, on traders' intuition, experience, skill, and luck. 
Moreover, currently, numerous online platforms and smartphone applications (e.g., Robinhood) allow people to trade directly from their smartphones. In these platforms, the data is presented graphically, and in most cases, the user decides and executes his trade upon the visual representation alone.
Therefore, it's reasonable to examine the usefulness of visual representations as input to the model.

\item The third contribution is that we show that the concept of visual time-series classification is effective and works on real data. A large fraction of the artificial-intelligence research is conceptual and works only on synthetic data. As will be shown, the concepts introduced in this paper are not only effective on real data, but they can be leveraged to deploy immediately as either a marketing recommendation tool and/or as a forecasting tool.
\end{enumerate}


\section{Data and Methods}\label{sec:data+methods}
In this study, we use Yahoo finance to analyze the daily values of all companies that contribute to the S\&P 500 index for the period 2010-2018 (hereafter SP500 data). These are large-cap companies that are actively traded on the American stock exchanges, and their capitalization covers the vast majority of the American equity market (e.g., \citeauthor{berk2013fundamentals}, \citeyear{berk2013fundamentals}). 
 
\begin{figure}[htb]
\centering
\noindent\includegraphics[width=1\columnwidth]{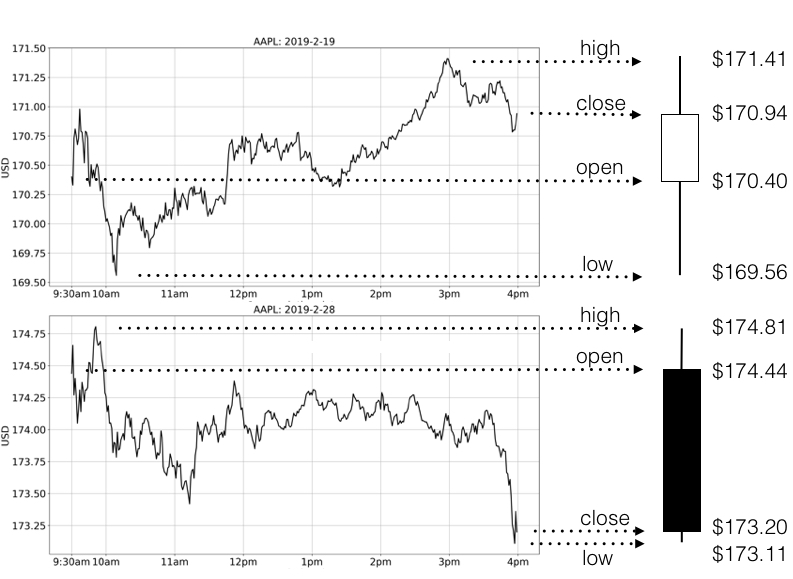}
\caption{Converting continuous time series to images. 
}
\end{figure}

Trading is done continuously (during trade hours which usually span between 9:30 am to 4:00 pm, not including before- and after-market hours). However, we use a discrete form of the continuous data by accounting only for the start, max, min, and end values per stock per day. These values are denoted, as is common in finance, as the Open, High, Low, and Close (OHLC) values (e.g., \citeauthor{murphy1999technical}, \citeyear{murphy1999technical}). We visualize the data using a box-and-whisker (also called candlestick) diagram, where box edges mark the Open and Close price, while the whiskers mark the Low and High values (i.e., daily min and max). The color of each box reveals whether the Open price finalized higher or lower than the Close price for the same day; if Open $>$ Close the box in filled in black indicating Bear's market, whereas if Open $<$ Close the box is filled in white indicating Bull's market (e.g., \citeauthor{murphy1999technical}, \citeyear{murphy1999technical}).
Figure 2 shows an example of this process by focusing attention on the AAPL ticker for Feb 19, 2019, and Feb 28, 2019. The left columns show the 1-minute continuous trading data during trading hours, while the right column detail the discretization process. Notice that the upper left time-series experiences a positive trend resulting in a white candlestick visualization, while the bottom left time-series data experiences a negative trend resulting in a black candlestick.

We compare three well-known binary indicators (\citeauthor{murphy1999technical}, \citeyear{murphy1999technical}), where each indicator is based on prescribed algebraic rules that depend solely on the Close values. Each indicator alerts the trader only for a buying opportunity. 
If a trader decides to follow (one of) the signals they may do so at any point no earlier than the day after the opportunity signal was created.
The three "buy" signals are defined as follows:
\begin{itemize}
\item BB crossing: 
The Bollinger Bands (BB) of a given time-series consists of two symmetric bands of 20-days moving two standard deviations (\citeauthor{colby1988encyclopedia}, \citeyear{colby1988encyclopedia}). The bands envelop the inherent stock volatility while filtering the noise in the price action. 
Traders use the price bands as bounds for trade activity around the price trend (\citeauthor{murphy1999technical}, \citeyear{murphy1999technical}).
Hence, when prices approach the lower band or go below, prices are considered to be in an oversold position and trigger a buying opportunity.
Here, the bands are computed using the (adjusted) Close values, and hence a buy signal is {\em defined} to trigger when the daily Close value {\em crosses above} the lower band.

Figure 3 shows an example of a Buy signal opportunities for the AAPL stock during 2018. 
In solid black, one can see the daily Close values for the ticker while the red line shows the 20-days moving average (inclusive) of the price line. The dashed black lines mark the two standard deviations above and below the moving average line. The BB crossing algorithm states that a Buy signal is initiated when the price line (in solid black) crosses above the lower dash black line. In this Figure, marked by the red triangles, one can identify eight such buy opportunities.

\begin{figure}[htb]
\centering
\noindent\includegraphics[width=1\columnwidth]{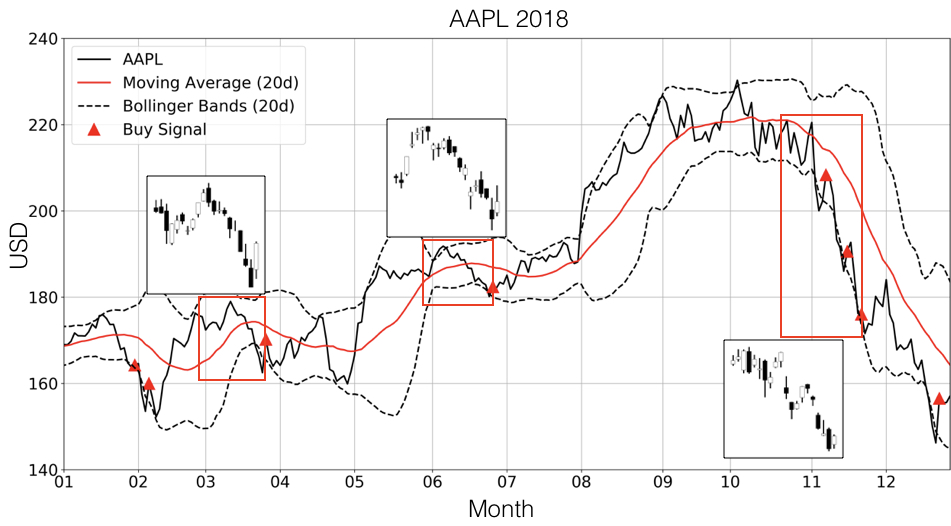}
\caption{Labeling time series data according to the Bollinger Bands crossing rule. 
}
\end{figure}

\item MACD crossing: 
Moving Average Convergence Divergence (MACD) is a trend-following momentum indicator that compares the relationship between short and long exponential moving averages (EMA) of an asset (\citeauthor{colby1988encyclopedia}, \citeyear{colby1988encyclopedia}). As is common in finance (e.g., \citeauthor{murphy1999technical}, \citeyear{murphy1999technical}), we compute the MACD by subtracting the 26-days EMA from the 12-days EMA. 
When MACD falls to negative values, it suggests negative momentum and conversely, when the MACD rises to positive values, it indicates upward momentum. Traders usually wait for consistent measures, thus smoothing the MACD line further by computing the 9-day EMA of the MACD, known as the signal line. 
Here, the MACD buy signal is {\em defined} to trigger when the signal line {\em crosses above}. 

\item RSI crossing: The Relative Strength Index (RSI) is an oscillating indicator that summarizes the magnitude of recent price changes to evaluate the overbought or oversold conditions of an asset. As is common in finance (e.g., 
\citeauthor{colby1988encyclopedia}, \citeyear{colby1988encyclopedia};
\citeauthor{murphy1999technical}, \citeyear{murphy1999technical}), we compute RSI as the ratio 14-days EMA of the incremental increase to the incremental decrease in asset values. The ratio is then scaled to values that vary between 0 and 100: it rises as the number and size of daily gains increases and falls as the number and size of daily losses increases. 
Traders use RSI as an indication of either an overbought or an oversold state. An overbought state might trigger a sell order; an oversold state might trigger a buy order. The standard thresholds for oversold/overbought RSI are 30/70, respectively (\citeauthor{murphy1999technical}, \citeyear{murphy1999technical}).
Here, the RSI buy signal is {\em defined} to trigger when the RSI line {\em crosses above} the value of RSI=30.
\end{itemize}

Figure 3 shows three positively-labeled images that correspond to the BB-crossing algorithm. These images are generated by enveloping 20 days of stock activity (the red rectangles) before and including the buy-signal day activity. It is also possible to create negatively-labeled images from this time-series by enveloping activity, in the same way, for days with no buy signal.
Note also that these images tightly bind the trade activity and do not contain labels, tickers, or title, which is the essential input data standardization pre-process we apply in this study.

\section{Results}\label{sec:results}
The objective of this study is to examine whether or not we can train a model to recover trade signals from algebraically-defined time-series data that is typical of technical analysis. We examine the supervised classification predictions of the time-series images that are labeled according to the BB, RSI, and MACD algorithms. 

The data set is balanced, containing 5,000 samples per class per indicator. That is, for each of the S\&P500 tickers, we compute all buy triggers for the period between 2010 and the end of 2017. We then choose, at random, 10 buy triggers for each ticker and create corresponding images. In the same way, we choose, at random, 10 no-buy triggers per ticker and create similar images. This process results in 10,000 high-resolution images per trigger.

A key difference between the three algorithms, besides their varying complexity, is the time-span each considers. While the BB algorithm takes into account 20 days of price action, RSI, which uses exponential-moving averaging considers (effectively) 27 days. MACD, which also uses exponential-moving averages, spans (effectively) over 26 days. For each of the triggers, we crop the images according to the number of effective trading days they consider. Thus, the BB images include information of 20 trade days, while RSI contains data for 27 days, and MACD, the most sophisticated algorithm that compares three time scales, contains 26 days of data.
In other words, each sample has 80-108 features depending on the size of the window required to compute the label (i.e., 4x20 for the BB crossing, and 4x26, 4x27 for the MACD and RSI respectively).

\begin{figure}[htb]
\centering
\noindent\includegraphics[width=1\columnwidth]{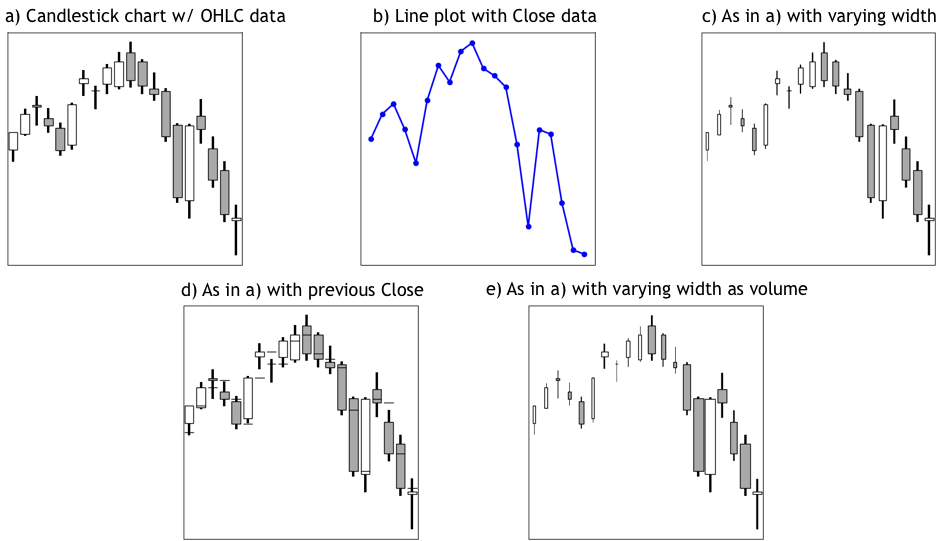}
\caption{Various visual representations of the same time-series data. 
}
\end{figure}

Figure 4 depicts an example of the five different visual designs we use in this study. Panel 4a uses the OHLC data as in Fig.~2, while panel 4b uses only the Close values plotted as a line chart. The design of 4b serves as a reference performance level, as will be discussed later.

In this study, a key element is how to express the direction and notion of time, or recency, in static images. 
A simple way of incorporating recency in the images is via the \emph{labels}. Each image is labeled according to trade opportunities, which are defined by crossing above a threshold at the right part of the image. The labels are time-dependent and are tied to a specific region in the chart; thus, implicitly, they deliver the notion of time to the static images.  
Another way of incorporating recency in the images is to incorporate the notion of time directly in them.
The designs at panels 4c and 4d aim at explicitly representing the direction of time by either linearly varying the width of the boxes towards the right (4c), or by overlaying the previous Close value as a horizontal line on each of the candlesticks (4d).
Lastly, in panel 4e, we augment the OHLC data by incorporating the trade volume in the candlestick visualization by varying the width of each box according to the relative change of the trade volume within the considered time frame. 
Remember that all three label-generating rules consider only the Close value, but each Close value is influenced by its preceding daily activity, reflected in the candlestick's visualization. We expect a trained model to either filter out unnecessary information or discover new feature relationships in the encoded image and identify the label-generating rule.

Following the above process, we create high-resolution images based on the discrete form of the data. Another question we have to address is what resolution do we need to maintain for proper analysis. The problem is that the higher the resolution, the more we amplify the (pixelated) feature space introducing more noise to the models and possibly creating unwanted spurious correlations.
We examine this point by varying the resolution of the input images in logarithmic scale and comparing the accuracy score of a hard voting classifier over the following 16 trained classifiers: Logistic Regression, Gaussian Naive-Bayes, Linear Discriminant Analysis, Quadratic Discriminant Analysis, Gaussian Process, K-Nearest Neighbors, Linear SVM, RBF SVM, Deep Neural Net, Decision Trees, Random Forest, Extra Randomized Forest, Ada Boost, Bagging, Gradient Boosting, and Convolutional Neural Net\footnote{The Deep Neural Net uses 32x32x32 structure, while the Convolutional Neural Net (CNN) uses three layers of 32 3x3 filters with ReLU activations and Max Pooling of 2x2 in between the layers. The last layer incorporates Sigmoid activation. The CCN model is compiled with Adam optimizer, binary-cross entropy loss function and run with a batch size of 16 samples for 50 iterations}. 
The focus here is on comparing the models' aggregated performance while changing the representation of the input space. 

\begin{figure}[htb]
\centering
\noindent\includegraphics[width=1\columnwidth]{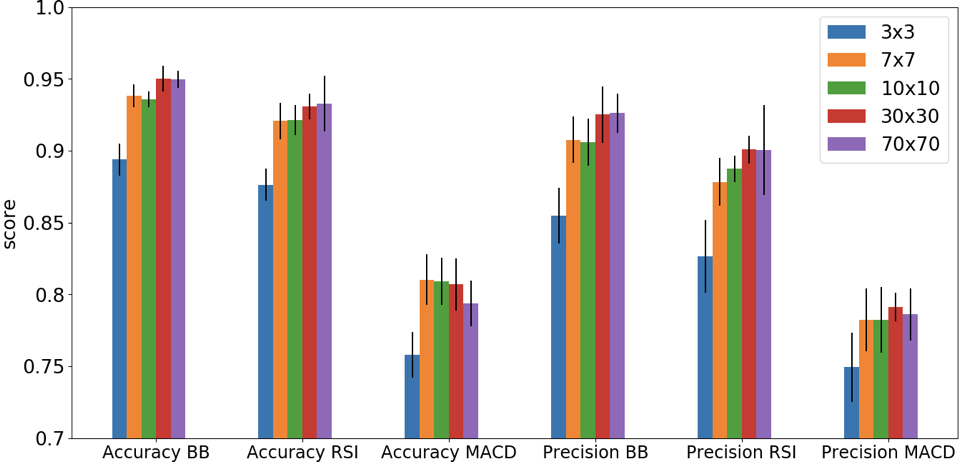}
\caption{The effect of varying the image resolution on the classification accuracy and precision scores for the three label-generating rules. 
}
\end{figure}

Figure 5 shows the results of the classification scores when we downscale the resolutions of the images that are labeled following the BB, RSI, and MACD algorithms. We use the Lanczos filter for downscaling, which uses sinc filters and efficiently reduces aliasing while preserving sharpness.
To evaluate the models' performances, we use the 5-fold cross-validation technique. This allows us to infer not only the mean prediction of the voting classifier but also the variability about the mean. (The vertical black lines in Fig.~5 show one symmetric standard deviation about the mean accuracy). Figure 5 shows that regardless of the labeling rule, the average accuracy and precision scores increase with finer resolutions but matures around 30x30 pixel resolution. For this reason, the following analysis is done using a 30x30 pixel resolution.

Figure 6 compares the predictability skill in the various image representations of the same input data for the three label-generating rules. All input representations perform remarkably well, and the predictability skill stands at about 95\% for the BB and RSI label-generating rules, while at approximately 80\% for the MACD labeled data. We were not surprised to see that the classifiers perform less efficiently on the MACD labeled data as this labeling-rule is the most complex involving multiple time-scales and smooth operations, all acting in concert.

\begin{figure}[htb]
\centering
\noindent\includegraphics[width=1\columnwidth]{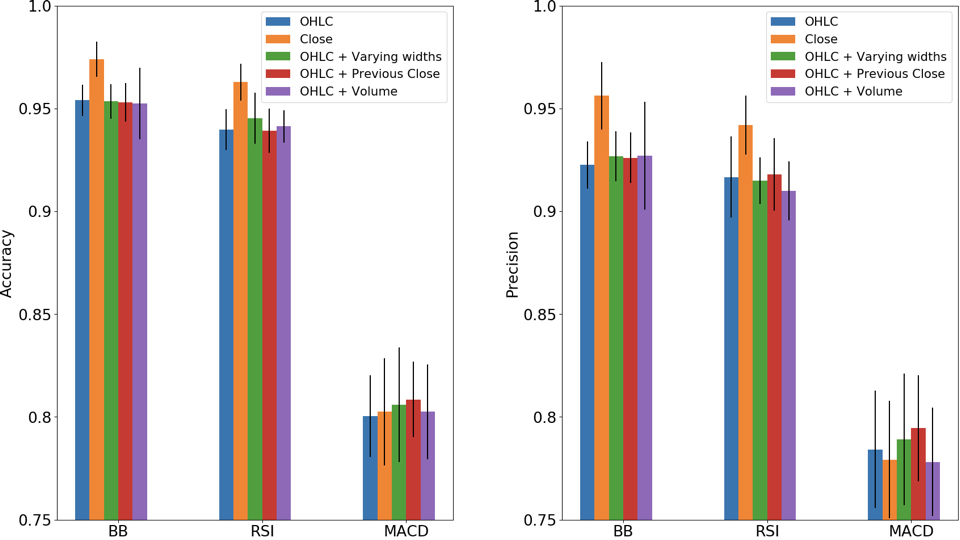}
\caption{The supervised classification accuracy (left panel) and precision (right panel) scores for the various triggers as a function of the different input representations.
}
\end{figure}

The best performing input data is the one that uses the Close values exclusively as line plots, while the various OHLC representations fall only a little behind. However, the line plot serves only as a point of reference -- the Bayesian performance level. This is because the label-generating rule depends exclusively on the Close values\footnote{Using the Close value alone is comparable to using the actual numerical data that the labeling rules are based upon.}. 
The key point is the fact that the various visual OHLC representations manage to achieve performance {\em comparable} to the Bayesian level. Most importantly, this finding is robust for the BB and RSI, as well as for the MACD algorithm.

Close examination of Fig.~6 shows that augmenting the OHLC input to include explicit time representation in the images by varying the bar widths linearly or by incorporating the previous Close values didn't add value. An exception to this is the MACD algorithm (a point we may explore further in a future study). On the other hand, encoding the irrelevant volume information in the candlestick images increased our uncertainty in predictions for all label-generating rules.
 
The precision score results are represented in the right panel of Fig.~6 and are almost identical to the accuracy scores on the left panel. 

\section{Discussion}\label{sec:discussion}
In this paper, we examine the supervised time-series classification task using large financial data sets and compare the results when the data is represented visually in various ways.
We find that even at low resolutions (see Fig.~5), time-series classification can be achieved effectively by transforming the task into a computer vision task. This finding is in accordance with \citeauthor{cohen2019b}, \citeyear{cohen2019b} who showed that classification of financial data using exclusive visual designs relates information spatially, aids in identifying new patterns and, in some cases, achieves {\em better} performance compared to using the raw tabular form of the same data.

Visualizing data and time-series data in particular, is an essential pre-processing step. Visualization by itself is not straightforward, especially for high-dimensional data, and it might take some time for the analyst to find the proper graphical design that will encapsulate the full complexity of the data. In this study, we essentially suggest considering the display as the input over the raw information. Our research indicates that even very complex multi-scale algebraic operations can be discovered by transferring the task to a computer vision problem.

A key question in this study is whether time-dependent signals can be detected in static images. To be more explicit, if the time axis goes left to right, it means that data points to the right are more recent and therefore, more critical to the model than data points to the left. But how can we convey this kind of information without an independent ordinal variable (e.g., time axis)?
We present two ways to incorporate the time-dependency in the images: the first leverages labels to deliver the notion of time; the second augments the images with sequential features. Incorporating time-dependency via labeling is done throughout the paper. We label the candlestick images using three algorithms and each computes a time-dependent function. Thus, each image encapsulates implicitly, via its corresponding label, the notion of time. That is, the signal we seek to detect is located on the right-most side of the image; the cross-above trigger always occurs because of the {\em last} few data points. In an example, the BB crossing algorithm effectively yields images with suspected local minimum on the right-hand side of the image. 
Incorporating time dependency explicitly by image augmentation is considered in two ways, by varying the width of the boxed in the candlestick diagram linearly and by overlaying the previous Close value on each candlestick.
It is noteworthy, however, that compared to the implicit label approach, we find the explicit augmentation to be less effective, as can be seen in Fig.~6.

In this study, we blended all S\&P 500 stocks and did not cluster the data by season, category, or sector. We used specific window sizes that correspond to the total length of information required by each algorithm to compute its label. 
To isolate the effect of the various window sizes, we examined the classification results when all window sizes were set to include 30 days of information. We found that the performance decreased when the window size added unnecessary information. (In this instance, there was a decrease in accuracy scores by a few percentage points -- not shown).

We see no need to account for the overall positive market performance during the 2010-2018 period as the analysis is done on a short times scales (about a month or less). One can complement this study by similarly analyzing for sell signals. We have repeated this analysis for sell signals and found that the overall quantitative results are very similar (not shown).

\begin{figure}[htb]
\centering
\noindent\includegraphics[width=1\columnwidth]{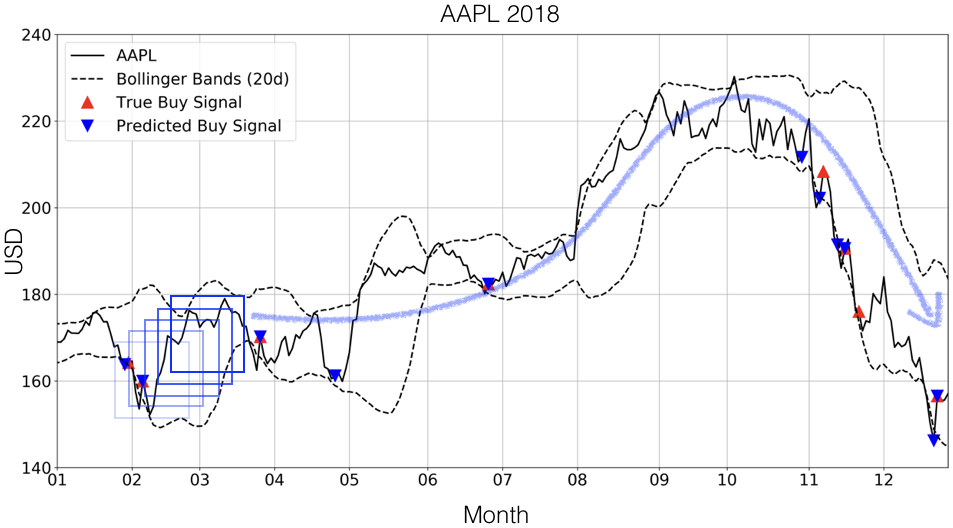}
\caption{Time-series forecasting using a 20-days rolling window. 
}
\end{figure}

We end this paper by noting that the supervised classification task can be applied as a forecasting tool (e.g.,
\citeauthor{hyndman2018forecasting}, \citeyear{hyndman2018forecasting}). In Fig.~7, we take out-of-sample, daily trading data from 2018. (As noted, the previous training and evaluation were computed using data from the period between 2010 and the end of 2017.) We create 20 days of images for \emph{every} day in the data. Next we feed these images to the voting classifier as a test set, and for each image, we predict what the label will be. Figure 7 corresponds to Fig.~3 but also includes blue triangles showing the predicted buy signal. Clearly, at least five buy signals were correctly classified, but even the missed ones are incredibly close in the sense that there is \emph{almost} cross-above the lower BB. Finally, depending on the use case, one can modify the binary probability threshold to achieve higher precision scores.

\section{Conclusion}\label{sec:conclusion}
Visual object recognition and object detection using machine learning and deep neural networks have shown great success in recent years (e.g.,
\citeauthor{krizhevsky2012imagenet}, \citeyear{krizhevsky2012imagenet};
\citeauthor{zeiler2014visualizing}, \citeyear{zeiler2014visualizing};
\citeauthor{szegedy2015going}, \citeyear{szegedy2015going};
\citeauthor{koch2015siamese}, \citeyear{koch2015siamese};
\citeauthor{lecun2015deep}, \citeyear{lecun2015deep};
\citeauthor{wang2017residual}, \citeyear{wang2017residual}).
In this paper, we follow up on these studies and examine the value in transforming numerical time-series analysis to that of image classification. We focus on financial trading after noticing that human traders always execute their trade orders while \emph{observing} images of financial time-series on their screens (see Fig.~1).  
Our study suggests that the transformation of time-series analysis to a computer vision task is beneficial for identifying trade decisions typical for humans using technical analysis.

\subsubsection*{Acknowledgments}
We would like to thank Jason Hunter, Joshua Younger, Alix Floman, and Veronica Bustamante for providing with insightful comments and crucial suggestions that helped in bringing this manuscript to completion. 

\subsubsection*{Disclaimer}
This paper was prepared for information purposes by the Artificial Intelligence Research group of J.~P.~Morgan Chase \& Co.~and its affiliates (“J.~P.~Morgan”), and is not a product of the Research Department of J.~P.~Morgan. J.~P.~Morgan makes no representation and warranty whatsoever and disclaims all liability, for the completeness, accuracy or reliability of the information contained herein.  This document is not intended as investment research or investment advice, or a recommendation, offer or solicitation for the purchase or sale of any security, financial instrument, financial product or service, or to be used in any way for evaluating the merits of participating in any transaction, and shall not constitute a solicitation under any jurisdiction or to any person, if such solicitation under such jurisdiction or to such person would be unlawful.   

\textcopyright 2020 J.~P.~Morgan Chase \& Co.~All rights reserved.


\bibliographystyle{ACM-Reference-Format}
\bibliography{sample-base}

\end{document}